\pdfoutput=1

\documentclass[11pt]{article}

\usepackage{coling}

\usepackage{times}
\usepackage{latexsym}

\usepackage[T1]{fontenc}

\usepackage[utf8]{inputenc}

\usepackage{microtype}

\usepackage{inconsolata}

\usepackage{graphicx}

\usepackage{enumitem}
\usepackage{multirow}
\usepackage{booktabs}
\usepackage{algorithm}
\usepackage{algorithmic}
\usepackage{tikz}
\usepackage{pgfplots}
\usepackage{graphicx}
\usepackage{subcaption}
\usepackage{pgf-pie}
\usepackage{enumitem}
\usepackage{pifont}
\usepackage{amsmath}
\usepackage{supertabular}
\usepackage{booktabs}
\numberwithin{equation}{section}

\usepackage{xcolor}
\definecolor{softred}{RGB}{250,100,100}
\definecolor{softgreen}{RGB}{56,118,29}
\definecolor{softblue}{RGB}{100,150,200}

\newcommand{\textblue}[1]{{\color{softblue}#1}}
\newcommand{\textgreen}[1]{{\color{softgreen}#1}}
\newcommand{\adamcolor}{pink}

\newcommand{\adam}[1]{\colorbox{\adamcolor}{$\displaystyle #1$}}
\newcommand{\adamtext}[1]{\colorbox{\adamcolor}{#1}}
\renewcommand{\algorithmiccomment}[1]{\bgroup\hfill $\triangleright$ ~#1\egroup}
\usepackage{bm}
\usepackage{xspace}
\newcommand{\ourmethod}{\textsc{Murre}\xspace}

\pgfplotsset{compat=1.18}
\definecolor{intro_red}{rgb}{0.651, 0.110, 0.0}
\definecolor{intro_green}{rgb}{0.153, 0.306, 0.075}
\newcommand{\introred}[1]{{\color{intro_red}#1}}
\newcommand{\introgreen}[1]{{\color{intro_green}#1}}
\definecolor{cpurple}{rgb}{0.675, 0.573, 0.922}
\definecolor{cblue}{rgb}{0.310, 0.757, 0.910}
\definecolor{cbluee}{rgb}{0.812, 0.886, 0.953}
\definecolor{cgreen}{rgb}{0.220, 0.463, 0.114}
\definecolor{corange}{rgb}{1, 0.808, 0.329}
\definecolor{cred}{rgb}{0.6, 0, 0}

\title{\ourmethod: Multi-Hop Table Retrieval with Removal\\for Open-Domain Text-to-SQL}

\author{
    Xuanliang Zhang, Dingzirui Wang, Longxu Dou, Qingfu Zhu, Wanxiang Che \\
    Harbin Institute of Technology \\
    \{xuanliangzhang, dzrwang, lxdou, qfzhu, car\}@ir.hit.edu.cn
}

\begin{document}
\maketitle
\begin{abstract}
The open-domain text-to-SQL task aims to retrieve question-relevant tables from massive databases and generate SQL.
However, the performance of current methods is constrained by single-hop retrieval, and existing multi-hop retrieval of open-domain question answering is not directly applicable due to the tendency to retrieve tables similar to the retrieved ones but irrelevant to the question.
Since the questions in text-to-SQL usually contain all required information, while previous multi-hop retrieval supplements the questions with retrieved documents.
Therefore, we propose the multi-hop table retrieval with removal (\ourmethod), which removes previously retrieved information from the question to guide the retriever towards unretrieved relevant tables.
Our experiments on two open-domain text-to-SQL datasets demonstrate an average improvement of $5.7\%$ over the previous state-of-the-art results.\footnote{Our code and data are available at \href{https://github.com/zhxlia/Murre}{https://github.com/zhxlia/Murre}.}

\end{abstract}

\section{Introduction}

Text-to-SQL, which simplifies database access and facilitates efficient data querying, is an important task in natural language processing \cite{qin2022surveytexttosql}.
Unlike the previous text-to-SQL task that provides the question-relevant database tables\footnote{For brevity, we refer to database tables relevant to the question as \textbf{relevant tables} in this paper.} \cite{yu-etal-2018-spider,shi2024survey_text-to-sql-llm}, a more realistic scenario involves open-domain text-to-SQL, where user questions are transformed into SQL across vast databases.
Therefore, open-domain text-to-SQL necessitates two main steps: retrieving relevant tables and generating SQL based on the retrieved tables and user question \cite{kothyari-etal-2023-crush4sql}.

\begin{figure}[t]
    \centering
    \includegraphics[width=1.0\linewidth]{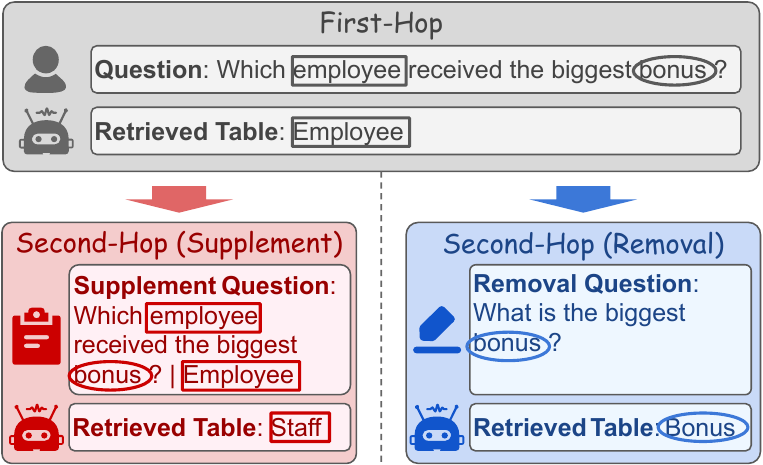}
    \caption{
        Comparison of multi-hop retrieval with supplementary and removal.
        The same shapes denote similar tables.
        Multi-hop retrieval with supplementary retrieves tables similar to those already retrieved, even if they are irrelevant to the question.
        In contrast, we employ multi-hop retrieval with removal which can successfully retrieve other relevant tables in the second hop.
    }
    \vspace{-1.5em}
    \label{fig:intro}
\end{figure}

To bridge the semantic gap between natural language user questions and structured database tables, CRUSH~\cite{kothyari-etal-2023-crush4sql} rewrites the user question into potentially relevant tables for retrieval. 
However, CRUSH employs single-hop retrieval, limiting its performance, as the retrieval of certain tables may depend on others, similar to multi-hop retrieval in open-domain question answering (QA) \cite{multi-hop-paragraph-retrieval,xiong2021Multi-Hop_Dense_Retrieval}. 
Nevertheless, directly adapting previous multi-hop retrieval methods could easily \textit{\textbf{retrieve tables that are similar to those retrieved in previous hops but irrelevant to the question}} since previous multi-hop retrieval usually supplements the retrieved documents into the user question \cite{shao-etal-2023-enhancing_iterretgen,rag_survey_2024,jeong2024adaptiverag}, as illustrated in Figure~\ref{fig:intro}.
However, open-domain text-to-SQL questions typically contain all necessary information.
For instance, in the left part of Figure~\ref{fig:intro}, the irrelevant table "\textit{staff}" is retrieved because it is similar to the "\textit{employee}" retrieved in the first hop.

To solve the above problem, we focus on exploring the multi-hop table retrieval method applicable to the open-domain text-to-SQL task, from the following two aspects:
(\emph{i})~We analyze why the multi-hop retrieval methods of open-domain QA cannot be directly applied to open-domain text-to-SQL.
(\emph{ii})~We propose a multi-hop table retrieval method based on removal for open-domain text-to-SQL, to ensure that the retrieved tables of each hop are relevant to the user question rather than just similar to the previously retrieved tables.

We first present that multi-hop retrieval of open-domain QA cannot be directly used for open-domain text-to-SQL.
We discuss that existing multi-hop retrieval methods are mainly to supplement the retrieved documents to the user question.
However, open-domain text-to-SQL needs to remove retrieved information rather than supplement it because all the necessary information is involved in the user question.
We conduct analysis experiments to prove our point of view.

Based on the above discussion, we propose a method called \textbf{MU}lti-hop table \textbf{R}etrieval with \textbf{RE}moval (\ourmethod), to enhance the performance of open-domain text-to-SQL through multi-hop retrieval.
As presented in the right part of Figure~\ref{fig:intro}, \ourmethod searches relevant tables employing multi-hop retrieval, by removing the retrieved table information from the question to ensure retrieved tables relevant to the user question.

To validate the effectiveness of \ourmethod, we conduct experiments on two datasets, SpiderUnion and BirdUnion, which are open-domain versions of text-to-SQL datasets Spider~\cite{yu-etal-2018-spider} and Bird~\cite{li2023bird}.
\ourmethod achieves an average improvement of $5.7\%$ compared to the previous state-of-the-art (SOTA) results, demonstrating its effectiveness.
Additionally, we calculate the average rank of relevant tables to demonstrate that \ourmethod can effectively retrieve relevant tables that are dissimilar to the previously retrieved tables, confirming that our method can indeed remove the retrieved information.

Our contributions are as follows:
\begin{itemize}[nolistsep,leftmargin=*]
    \item To enhance the retrieval performance in open-domain text-to-SQL, we discuss why multi-hop retrieval in open-domain QA can not be directly applied to text-to-SQL.
    \item To ensure that the retrieved tables are relevant to the user question, we present \ourmethod, prompting LLMs to remove the retrieved information from the question at each hop.
    \item To demonstrate the effectiveness of \ourmethod, we conduct experiments on the SpiderUnion and BirdUnion datasets, achieving an average improvement of $5.7\%$ compared with the previous SOTA results, proving its effectiveness.
\end{itemize}

\section{Analysis}
    \label{sec:analysis}
In this section, we present that \textbf{the multi-hop retrieval of open-domain QA cannot be directly applied to the open-domain text-to-SQL}.
We first discuss the multi-hop retrieval methods in open-domain QA.
Based on the above discussion, we analyze and conduct experiments to show the inapplicability of existing multi-hop retrieval.

\subsection{Multi-Hop Retrieval of Open-Domain QA}
    Multi-hop retrieval refers to retrieving with multiple hops, where each hop uses the information that is retrieved in previous hops \cite{xiong2021Multi-Hop_Dense_Retrieval,lee-etal-2022-generative_multi-hop}.
    Current multi-hop retrieval supplements the question with the retrieved information, guiding to identify the documents relevant to the currently retrieved documents \cite{rag_survey_2024,retrieval_survey,press-etal-2023-self_ask,li2023llatrieval,jeong2024adaptiverag}, since the user question could not contain all the information needed to retrieve all relevant documents. 
    For example, for the question \textit{"Ralph Hefferline was a psychology professor at a university that is located in what city?"}, we should first retrieve the university where he works, supplement it into the question, and then search for the city where the university is located.
    
\subsection{Previous Multi-Hop Retrieval is Unsuitable for Open-Domain Text-to-SQL}
    Unlike open-domain QA, the questions in open-domain text-to-SQL do not need retrieved information to supplement because they usually contain all the required information.
    For instance, in Figure~\ref{fig:intro}, the question \textit{"Which employee received the biggest bonus?"} contains all the relevant table names.
    Therefore, we propose to remove the retrieved information from the question, to retrieve the unretrieved relevant tables.
    Table~\ref{tab:ablation} presents that multi-hop retrieval with supplementary even underperforms single-hop retrieval, proving the necessity of removing retrieved information.

\begin{figure*}
    \centering
    \includegraphics[width=0.9\linewidth]{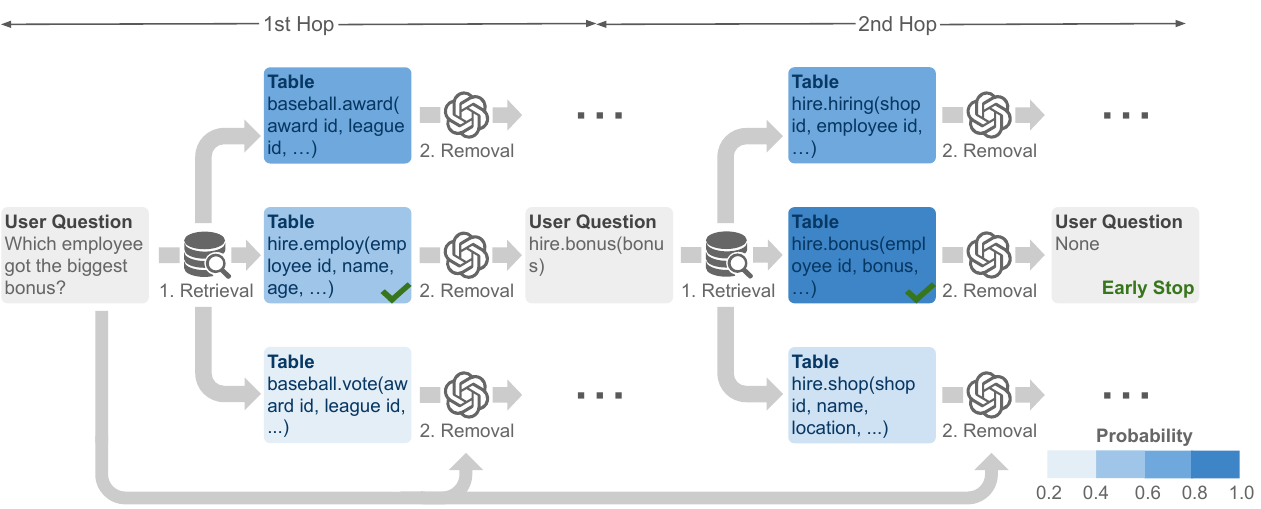}
    \caption{
        An overview of \ourmethod. Each hop consists of: 
        (\emph{i})~\textbf{Retrieval}: retrieving tables similar to the question; 
        (\emph{ii})~\textbf{Removal}: removing the retrieved information from the user question and representing the unretrieved information in the tabular format with LLM.
        We employ the beam search paradigm maintaining multiple retrievals at each hop.
        The \textblue{color} depth represents the probability that the table is relevant to the question of the hop, and \textgreen{\ding{51}} denotes the relevant table.
        We demonstrate an example that early stops at the second hop for brevity.
    }
    \label{fig:method}
\end{figure*}

\section{Methodology}
    \label{sec:methodology}
    \subsection{Task Definition}
    Our work mainly focuses on the open-domain text-to-SQL task, which can be formally defined as:
    Given a user question $q$, database tables $T=\{t_i\}$ and a number of retrieved tables $N$, suppose the tables relevant to $q$ are $T^q = \{t^q_i\}$, \ourmethod aims to retrieve $N$ tables $T^{q}_N$, where $T^q \subseteq T^{q}_N$.

\subsection{Overview}
    The overview of \ourmethod is illustrated in Figure~\ref{fig:method}.
    To reduce the error cascades in multi-hop retrieval, we adopt the beam search paradigm inspired by \citet{zhang2023beam_retrieval}.
    In \ourmethod, each hop can be divided into two phases: \textbf{Retrieval} (\S\ref{subsec:retrieve}) and \textbf{Removal} (\S\ref{subsec:rewrite}). 
    In each hop, we first \textbf{retrieve} the relevant tables by calculating their probability that is relevant to the question. 
    To avoid retrieving similar to the previous retrieved tables but irrelevant tables, we \textbf{remove} the retrieved information from the user question using LLM. 
    \ourmethod repeats the Retrieval and Removal phases until reaching the maximum hop limit $H$ or meeting the early stop condition.
    After the multi-hop retrieval, we \textbf{score} (\S\ref{subsec:rank}) each table based on its probability that is relevant to the user question and select the top-$N$ tables as the input for generating SQL. 
    In Appendix~\ref{appendix:no relevant tables}, we explore how to handle the situation when no table in the database is relevant to the user question.

\subsection{Retrieval}
\label{subsec:retrieve}
    The Retrieval phase aims to identify $B$ tables corresponding to the user question $q^{h,b}$, where $B$ is the beam size and $q^{h,b}$ denotes the user question at hop $h$ and beam $b$. 
    We embed the user question and each table into vectors and then compute their relevance probability to the user question. 
    Let \texttt{Emb}($x$) represent the embedding vector of $x$, the probability is expressed as Equation~\ref{equ:p_hat}. 
    \begin{equation}
        \hat{P}(t_i | q^{h,b}) = \texttt{Norm}(\frac{\texttt{Emb}(t_i) \cdot \texttt{Emb}(q^{h,b})}{|\texttt{Emb}(t_i)||\texttt{Emb}(q^{h,b})|})
        \label{equ:p_hat}
    \end{equation}
    We compute the probability using the cosine similarity between the question and table vectors, applying the $\texttt{Norm}$ function to ensure the conditional probabilities sum to one. 
    The detailed representations of the table and the normalization method are provided in Appendix~\ref{appendix:prompts for rewrite} and Appendix~\ref{appendix:normalization method}, respectively. 
    We then select the top $B$ tables with the highest probabilities as the retrieval results for the current hop question $q^{h,b}$.

    In the previous hop of retrieval, there are $B$ beams and each beam corresponds to $B$ retrieved tables obtained at the current hop, resulting in a total of $B \times B$ retrieval results.
    Following the beam search paradigm, we then choose $B$ results from these for the subsequent Removal phase, with the selection method detailed in \S\ref{subsec:rank}.

\subsection{Removal}
\label{subsec:rewrite}
    The Removal phase is designed to mitigate the retrieval of tables that are similar to previously retrieved ones yet irrelevant to the user question.
    To achieve the above goal, this phase prompts LLMs to remove the information of retrieved tables in $\texttt{Path}^{h,b}$ (which includes all retrieved tables from hop $1$ to hop $h$ on the path of beam $b$) from the user question $q$, and express the unretrieved information in the form of a table with some demonstrations, following the previous work \cite{kothyari-etal-2023-crush4sql}. 
    The removal question is used to guide the retriever for the next hop. 

    Considering that different user questions require varying numbers of tables, to prevent additional hops from introducing errors, we instruct LLMs to assess whether the retrieved tables in $\texttt{Path}^{h,b}$ are sufficient to answer $q$, i.e., early stop. 
    In detail, we prompt the LLM to generate a special mark when removing the retrieved information (e.g., "None" in Figure~\ref{fig:method}) to indicate that the retrieved tables are sufficient to answer the user question $q$, ceasing further retrieval as soon as this early stop mark is produced. 
    The prompts used for removal are presented in Appendix~\ref{appendix:prompts for rewrite}.


\subsection{Score}
\label{subsec:rank}
    The aforementioned multi-hop retrieval process maintains $B$ paths at each hop, containing tables that could exceed or fall short of the required number of tables $N$. 
    Therefore, this phase aims to score all the tables based on their probabilities to retain the most relevant tables used for generating SQL, which consists of two parts: 
    (\emph{i}) scoring the retrieval path $\texttt{Score\_Path}(\texttt{Path}^{h,b})$;
    (\emph{ii}) scoring the retrieved table $\texttt{Score\_Table}(t_i)$.
    
    \paragraph{Retrieval Path Score}
        First, we address the calculation of the retrieval path score, where each node on the path corresponds to the retrieved table on a beam of one hop. 
        As discussed in \S\ref{subsec:retrieve}, the score of each retrieval path represents the probability that the last table in the path is retrieved, given its corresponding question at the last hop. 
        Following the derivation in Appendix~\ref{appendix:score of the retrieval path}, the score of a retrieval path is computed as the product of all the probabilities $\hat{P}$ in the path.

    \paragraph{Table Score}
        Building on the retrieval path score, we describe the calculation to score a retrieved table $t_i$. 
        Since each table could have multiple scores across various hops and beam retrievals, we propose a table scoring algorithm to effectively integrate these scores.
        Considering the potential interrelation of question-relevant tables, we aim to ensure that all retrieved tables collectively are the most relevant to the question.
        Thus, the higher the retrieval path score, the higher the score of the tables in the path. 
        Specifically, let $\texttt{Path}_{t_i}$ denote all retrieval paths containing table $t_i$, we calculate the table score as: $\texttt{Score\_Table}(t_i) = \mathop{\max}_{t \in Path_{t_i}} \texttt{Score\_Path}(t)$. 
        Finally, we select $T^Q_N = \{t_1, ..., t_N\}$ with the highest $\texttt{Score\_Table}(t_i)$ as our retrieval results. 
        The detailed algorithm is provided in Appendix~\ref{appendix:table scoring algorithm}. 
        Appendix~\ref{appendix:synonyms} elaborates on enhancing the table scoring algorithm to address ambiguous entities or synonyms in user questions or tables.

\section{Experiments}
    \label{sec:experiments}
        
    \subsection{Experiment Setup} 
    \begin{table}[t]
        \centering
        \small
        \begin{tabular}{l|ccccc}
            \toprule
            \multirow{2}{*}{\textbf{Dataset}} & \multicolumn{5}{|c}{\textbf{\#Table}}\\
             & \bm{$1$} & \bm{$2$} & \bm{$3$} & \bm{$4$} & \textbf{All} \\
            \midrule
            SpiderUnion & $395$ & $214$ & $43$ & $6$ & $658$\\
            BirdUnion & $364$ & $943$ & $207$ & $20$ & $1534$\\ 
            \bottomrule
        \end{tabular}
        \caption{
            The distribution of questions based on the number of relevant tables (\textbf{\#Table}) on the SpiderUnion and BirdUnion.
            \textbf{All} refers to the total number of questions in the dataset.
        }
        \label{tab:dataset}
    \end{table}

        \begin{table*}[t]
        \centering
        \small
        \begin{tabular}{lll|cccc|cccc}
            \toprule
            \textbf{Dataset} & \textbf{Model} & \textbf{Method} & \bm{$k=3$} &  \bm{$k=5$} &  \bm{$k=10$} &  \bm{$k=20$} & \bm{$r@3$} & \bm{$r@5$} & \bm{$r@10$} & \bm{$r@20$}\\
            \midrule
            \multirow{6}{*}{SpiderUnion} & \multirow{3}*{SGPT-125M} & Single-hop &  $54.3$ & $66.0$ & $75.4$ & $82.2$ & $63.0$ & $73.1$ & $80.7$ & $86.3$ \\
            ~ & ~ & CRUSH$^\dag$ & $60.2$ & $71.3$ & $80.7$ & \bm{$86.8$} &  $68.9$ & $76.3$ & \bm{$83.4$} & \bm{$88.9$}\\
            ~ & ~ & \ourmethod &  \bm{$65.0$} & \bm{$74.2$} & \bm{$81.0$} & $85.3$ & \bm{$70.2$} & \bm{$77.5$} & $82.3$ & $86.9$\\
            \cmidrule{2-11}
            ~ &  \multirow{3}{*}{SGPT-5.8B} & Single-hop &  $76.3$ & $86.8$ & $94.1$ & \bm{$97.6$} & $84.0$ & $91.5$ & $96.2$ & \bm{$98.7$}\\
            ~ & ~ & CRUSH$^\dag$ & $68.2$ &  $80.1$ & $88.4$ & $92.2$ & $75.5$ & $85.1$ & $91.2$ & $94.5$ \\
            ~ & ~ & \ourmethod & \bm{$86.0$} & \bm{$93.5$} & \bm{$96.7$} & $97.3$ & \bm{$89.3$} & \bm{$94.3$} & \bm{$96.8$} & $97.5$\\
            \midrule
            \multirow{6}{*}{BirdUnion} & \multirow{3}*{SGPT-125M} & Single-hop & $39.0$ & $50.3$ & $62.1$ & $70.9$ & $54.0$ & $63.2$ & $73.3$ & $80.9$ \\
            ~ & ~ & CRUSH$^\dag$ & $42.1$ & $56.1$ & $70.2$ & $77.7$ & $60.2$ & $70.0$ & $79.5$ & \bm{$86.1$} \\
            ~ & ~ & \ourmethod & \bm{$51.4$} & \bm{$62.7$} & \bm{$72.9$} & \bm{$78.3$} & \bm{$64.8$} & \bm{$72.7$} & \bm{$79.6$} & $84.2$\\
            \cmidrule{2-11}
            ~ &  \multirow{3}{*}{SGPT-5.8B} & Single-hop & $55.3$ & $67.3$ & $79.4$ & $86.4$ & $72.9$ & $80.8$ & $88.6$ & $92.8$ \\
            ~ & ~ & CRUSH$^\dag$ & $52.2$ &  $63.5$ & $78.4$ & $88.1$ & $70.0$ & $77.9$ & $87.5$ & $93.0$ \\
            ~ & ~ & \ourmethod & \bm{$69.1$} & \bm{$80.1$} & \bm{$88.7$} & \bm{$92.7$} & \bm{$81.0$} & \bm{$87.6$} & \bm{$92.6$} & \bm{$95.4$}\\
            \bottomrule
        \end{tabular}
        \caption{
            The complete recall ($k$) and recall ($r$) of \ourmethod, compared with Single-hop and CRUSH on SpiderUnion and BirdUnion, using SGPT-125M and SGPT-5.8B as the embedding models. 
            $^\dag$ denotes our run since the performance difference led by the API change.
            The best results of each dataset and model are annotated in \textbf{bold}.
        }
        \label{tab:main result}
    \end{table*}


    \paragraph{Dataset}
        To evaluate the effectiveness of \ourmethod, we validate it on two open-domain text-to-SQL datasets: SpiderUnion~\cite{kothyari-etal-2023-crush4sql} and BirdUnion, which combine the tables of Spider~\cite{yu-etal-2018-spider} and Bird~\cite{li2023bird}, following previous works~\cite{kothyari-etal-2023-crush4sql,chen2024table_retrieval}.
        Spider is a widely used cross-domain dataset for the text-to-SQL task, while Bird is more reflective of real-world scenarios, featuring more complicated queries.
        Table~\ref{tab:dataset} presents the distribution of questions based on the number of tables required. 
        Further details on Spider and Bird datasets are provided in Appendix~\ref{appendix:dataset details}.

    \paragraph{Metric}
        We employ recall and complete recall as evaluation metrics for retrieval, and Execution Accuracy (EX) \cite{yu-etal-2018-spider} for text-to-SQL. 
        Recall (\bm{$r@$}) measures the proportion of relevant tables retrieved from all relevant tables, following previous work \cite{kothyari-etal-2023-crush4sql}.
        Unlike other open-domain tasks (e.g., open-domain QA), retrieving all relevant tables is critical for generating correct SQL in open-domain text-to-SQL. 
        Therefore, we introduce complete recall (\bm{$k=$}), which is the proportion of examples where all relevant tables are retrieved.
        For text-to-SQL, we follow \cite{gao2023texttosqldail} in using Execution Accuracy (EX) to assess the correctness of execution results by comparing the predicted SQL to the gold standard. 
    
    \paragraph{Model}
        We utilize SGPT~\cite{muennighoff2022sgpt} to embed tables and user questions without additional fine-tuning, following the previous work~\cite{kothyari-etal-2023-crush4sql}.
        For the Removal phase and SQL generation, we use \texttt{gpt-3.5-turbo}\footnote{\href{https://platform.openai.com/docs/models/gpt-3-5}{Document} for \texttt{gpt-3.5-turbo}} to predict.  
        The detailed descriptions of SGPT and \texttt{gpt-3.5-turbo} are provided in Appendix~\ref{appendix:model details}.

    \paragraph{Comparing System}
        In our experiments, we compare \ourmethod with the following methods: 
        (\emph{i}) Single-hop, which retrieves tables based on the user question in a single hop;
        (\emph{ii}) CRUSH~\cite{kothyari-etal-2023-crush4sql}, which retrieves tables in a single hop by converting the user question into a table format through hallucination.

    \paragraph{Implement Details}
        We set the beam size to $5$, as it provides the best performance with the smallest size (see \S\ref{subsubsec:beam size}).
        The maximum hop (abbreviated as max hop) is set to $3$ because over $98\%$ of questions in SpiderUnion and BirdUnion require $\leq 3$ tables (see Table~\ref{tab:dataset}).
        We use a $9$-shot prompt for SpiderUnion and an $8$-shot prompt for BirdUnion to remove the retrieved information, given the larger table scales in BirdUnion compared to SpiderUnion.

\begin{table*}[htbp]
    \centering
    \small
    \begin{tabular}{ll|cccc|cccc}
        \toprule
        \multirow{2}{*}{\textbf{Model}} & \multirow{2}{*}{\textbf{Method}} & \multicolumn{4}{c|}{\textbf{SpiderUnion}} & \multicolumn{4}{c}{\textbf{BirdUnion}} \\
        & & \bm{$r@3$} &  \bm{$r@5$} & \bm{$r@10$} & \bm{$r@20$} & \bm{$r@3$} &  \bm{$r@5$} & \bm{$r@10$} & \bm{$r@20$}\\
        \midrule
        \multirow{3}{*}{SGPT-125M} & Single-hop & $43.9$ & $50.0$ & $53.2$ & $54.1$ & $11.2$ & $13.4$ & $17.1$ & $18.5$ \\
        ~ & CRUSH & $47.3$ & $50.9$ & \bm{$55.9$} & \bm{$59.6$} & $14.5$ & \bm{$17.1$} & $19.0$ & \bm{$20.5$} \\
        ~ & \ourmethod & \bm{$50.3$} & \bm{$54.1$} & $54.7$ & $57.4$ & \bm{$16.0$} & $16.8$ & \bm{$19.4$} & $20.0$ \\
        \cmidrule{1-10}
        \multirow{3}{*}{SGPT-5.8B} & Single-hop & $54.9$ & $60.6$ & $61.6$ & $63.7$ & $16.9$ & $17.3$ & $18.3$ & $20.4$ \\
        ~ & CRUSH & $49.8$ & $56.4$ & $60.3$ & $60.8$ & $16.8$ & $18.0$ & $19.7$ & $21.1$ \\
        ~ & \ourmethod & \bm{$62.5$} & \bm{$64.4$} & \bm{$64.4$} & \bm{$66.7$} & \bm{$20.9$} & \bm{$21.8$} & \bm{$22.0$} & \bm{$22.4$} \\
        \bottomrule
    \end{tabular}
    \caption{
        EX for predicted SQL based on the user question and varying numbers of retrieved tables. 
    }
    \label{tab:text-to-sql}
\end{table*}

\subsection{Main Result} 
    The main results of our experiments are presented in Table~\ref{tab:main result}. 
    Compared to CRUSH, \ourmethod demonstrates significant improvements across datasets and models of different scales, with an average increase of $5.7\%$ in recall and complete recall over the previous SOTA, which validates the effectiveness of our method. 
    We can also see that:

    \paragraph{The improvement of \ourmethod on BirdUnion is more significant than on SpiderUnion.}
        Since the questions in BirdUnion typically require more tables (see Table~\ref{tab:dataset}), requiring multi-hop retrieval of \ourmethod more to obtain multiple relevant tables, thereby enhancing retrieval performance.
        
    
    \paragraph{As the number of top-ranked tables grows, the retrieval performance of \ourmethod slows down.}
        Improving metrics with a high number of top-ranked tables necessitates retrieving relevant tables that are highly dissimilar to the user question, making the metrics challenging to enhance.
        Especially, for some metrics (e.g., $k=20$, $r@20$), the performance of \ourmethod declines slightly because removing retrieved information shifts the focus toward retrieving increasingly dissimilar tables.

    \paragraph{Text-to-SQL Experiments}
        We conduct text-to-SQL experiments on SpiderUnion and BirdUnion using the user question and retrieved tables as the input, as shown in Table~\ref{tab:text-to-sql}. 
        The performance of \ourmethod exceeds that of both Single-hop and CRUSH, 
        validating the effectiveness of our method.
        As the number of input tables increases, EX improvements slow beyond $10$ tables because too many irrelevant tables hinder the model from focusing on the relevant ones. 
        This also underscores the necessity of \ourmethod in enhancing the retrieval performance with a limited set of top-ranked tables in the open-domain text-to-SQL task.
    
\subsection{Ablation Studies}
\label{subsec:ablation}
    \begin{table*}[htpb]
        \centering
        \small
        \begin{tabular}{l|ccc|ccc}
            \toprule
            \textbf{Method} & \bm{$k=3$} &  \bm{$k=5$} &  \bm{$k=10$}  & \bm{$r@3$} & \bm{$r@5$} & \bm{$r@10$} \\
            \midrule
            \ourmethod & \bm{$65.0$} & \bm{$74.2$} & \bm{$81.0$} & \bm{$70.2$} & \bm{$77.5$} & \bm{$82.3$}\\
            \textit{w/o removal} & $46.2$ ($-18.8$) & $56.7$ ($-17.5$) & $67.2$ ($-13.8$) & $50.6$ ($-19.6$) & $60.7$ ($-16.8$) & $70.0$ ($-11.6$) \\
            \textit{w/o tabulation} & $54.6$ ($-10.4$) & $64.9$ ($-9.3$) & $75.5$ ($-5.5$) & $63.4$ ($-6.8$) & $72.5$ ($-5.0$) & $80.9$ ($-1.4$) \\
            \textit{w/o early stop} & $52.6$ ($-12.4$) & $64.9$ ($-9.3$) & $71.0$ ($-10.0$) & $57.1$ ($-13.1$) & $67.0$ ($-10.5$) & $72.2$ ($-10.1$) \\
            \bottomrule
        \end{tabular}
        \caption{
            The ablation results on evaluating \ourmethod, compared with splicing the question with retrieved tables (denoted as \textit{w/o removal}), querying the rest information with natural language after removing retrieved information (denoted as \textit{w/o tabulation}), and without employing the mechanism of early stop (denoted as \textit{w/o early stop}) on SpiderUnion with SGPT-125M. 
        }
        \label{tab:ablation}
    \end{table*}
    To demonstrate the effectiveness of our method, we conduct ablation experiments on SpiderUnion, with results presented in Table~\ref{tab:ablation}.
    We select SpiderUnion corresponding to Spider for subsequent experiments because Spider is the mainstream text-to-SQL dataset.
    Since SGPT-125M and SGPT-5.8B exhibit similar trends across datasets and methods (see Table~\ref{tab:main result} and \ref{tab:text-to-sql}), we select SGPT-125M as the embedding model employed for subsequent experiments, considering both the embedding speed and the retrieval recall \cite{muennighoff-etal-2023-mteb_Text_Embedding_Benchmark}.

    \paragraph{The Effectiveness of Removal}
        To demonstrate the effectiveness of Removal which removes retrieved information from the user question, we compare its performance against the popular multi-hop retrieval method in open-domain QA, which directly splices the user question with retrieved tables at each hop. 
        Compared with \ourmethod, the performance of the splicing method drops significantly, underscoring the effectiveness of Removal which alleviates retrieving similar but irrelevant tables.

    \paragraph{The Effectiveness of Tabulation}
        To prove the effectiveness of transforming questions into a tabular format (abbreviated as tabulation), we query the removed questions with natural language.
        The results indicate that, compared to querying with natural language, tabulation significantly enhances performance, validating its effectiveness in \ourmethod. 
        The prompt without tabulation is presented in Appendix~\ref{appendix:prompts for rewrite}. 

    \paragraph{The Effectiveness of Early Stop}
        To verify the effectiveness of early stop in \ourmethod, we compare the results without using this mechanism, where the model does not generate the special early stop mark.
        The performance without the early stop shows a significant decline, demonstrating that its inclusion in \ourmethod is crucial for maintaining performance.
        
\subsection{Analysis}
    In the analysis experiments, 
    we use $k=5$ as the evaluation metric, with detailed explanations in Appendix~\ref{appendix:the evaluation metric used in analysis experiments}. 
    Additionally, we discuss the efficiency of \ourmethod in Appendix~\ref{appendix:discussion on efficiency}, the reasons we use the original user question at the first hop in Appendix~\ref{app:why not rewrite}, and the impact of SQL hardness in Appendix~\ref{appendix:impact of sql hardness}.

    \subsection{Why \ourmethod improve compared with Single-hop method?}
        \begin{figure}
            \centering
            \resizebox{0.75\linewidth}{!}{
    \begin{tikzpicture}[scale=0.6]
        \small
        \pie[
            pos={8,0},
            color={cpurple!50, cblue!30, white!20},
            text=legend
        ]{
            87.0/Similar Candidate ($80$),
            14.9/Semantic Gap ($14$)
        }
    \end{tikzpicture}
}
            \caption{
                The proportion of performance improvements achieved by \ourmethod in addressing various limitations, compared to Single-hop.
                The numbers in parentheses in the legend represent the number of improved examples.
            }
            \label{fig:improvement}
        \end{figure}
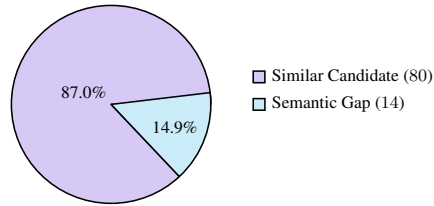
        To explore why \ourmethod enhances retrieval performance, we analyze its improvement compared with Single-hop in Figure~\ref{fig:improvement}.
        We observe that: 
        (\emph{i})~Similar Candidate means that many irrelevant candidate tables are similar to the question.
        \ourmethod alleviates the limitation by removing the retrieved information, allowing the model to focus on retrieving tables not yet retrieved. 
        (\emph{ii})~Semantic Gap refers to the semantic distinction between the natural language question and the relevant tables.
        \ourmethod tabulates the question concerning the retrieved information, aligning the question more closely with tables in terms of domain and vocabulary, thus narrowing the semantic gap \cite{knowledge-injection,li2024enhancingllmfactualaccuracy}. 
        Detailed examples illustrating the advantages of \ourmethod are provided in Appendix~\ref{appendix:detailed case study}, with the statistical criteria in Appendix~\ref{appendix:statistical criteria of limitations}.

    \subsubsection{How Does Beam Size Affect the Performance?}
    \label{subsubsec:beam size}
         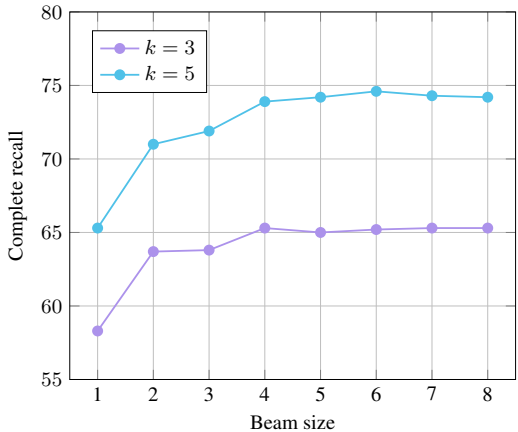
\begin{figure}
            \centering
        
        

\resizebox{0.9\linewidth}{!}{
    \begin{tikzpicture} 
        \small
        \begin{axis}[
            xlabel=Beam size,
            ylabel=Complete recall,
            xmin=-0.5, xmax=7.5, 
            ymin=55, ymax=80,
            xtick={0,1,2,3,4,5,6,7},
            xticklabels={1,2,3,4,5,6,7,8},
            grid=both, 
            legend style={
                at={(0.05,0.95)}, 
                anchor=north west, 
                font=\small, 
                legend cell align=left
            }
            ]
        \addplot[cpurple, line width = 0.8, mark = *] plot coordinates {
            (0,58.3)
            (1,63.7)
            (2,63.8)
            (3,65.3)
            (4,65.0)
            (5,65.2)
            (6,65.3)
            (7,65.3)
        };
        \addlegendentry{$k=3$}
        
        \addplot[cblue, line width = 0.8, mark = *] plot coordinates {
            (0,65.3)
            (1,71.0)
            (2,71.9)
            (3,73.9)
            (4,74.2)
            (5,74.6)
            (6,74.3)
            (7,74.2)
        };
        \addlegendentry{$k=5$}
        
        \end{axis}
    \end{tikzpicture}
}
            \caption{
                The complete recall with different beam sizes on SpiderUnion with SGPT-125M.
            }
            \vspace{-1em}
            \label{fig:beam}
        \end{figure}
        
        To observe the impact of different beam sizes on the retrieval performance, we compare the performance of \ourmethod with different beam sizes, as shown in Figure~\ref{fig:beam}.
        (\emph{i})~When \ourmethod does not employ beam search, i.e., with a beam size of $1$, performance degrades rapidly, suggesting that beam search mitigates the effects of error cascade.
        (\emph{ii})~As the beam size increases, complete recall improves significantly until a beam size of $5$, after which performance either plateaus or declines. 
        This indicates that while employing beam search enhances performance, a beam size $\leq 5$ introduces too many irrelevant tables, increasing computational cost without further performance improvement.

    \subsubsection{How Does the Number of Hops Affect the Performance?}
        \begin{table}[t]
            \centering
            \small
            \begin{tabular}{c|ccccc}
                \toprule
                \multirow{2}{*}{\textbf{Max Hop}} & \multicolumn{5}{|c}{\textbf{\#Table}}\\
                 & \bm{$1$} & \bm{$2$} & \bm{$3$} & \bm{$\geq 4$} & \textbf{All} \\
                \midrule
                $1$ & $73.7$ & $59.8$ & $25.6$ & \bm{$50.0$} & $66.0$\\
                $2$ & $73.2$ & $77.6$ & \bm{$58.1$} & \bm{$50.0$} & $73.4$\\
                $3$ & \bm{$74.2$} & \bm{$78.0$} & \bm{$58.1$} & \bm{$50.0$} & \bm{$74.2$}\\
                $4$ & \bm{$74.2$} & \bm{$78.0$} & \bm{$58.1$} & \bm{$50.0$} & \bm{$74.2$}\\
                \bottomrule
            \end{tabular}
            \caption{
                Complete recall $k=5$ of \ourmethod with varying max hops. 
                SpiderUnion is divided based on the number of relevant tables per question (denoted as \textbf{\#Table}). 
                \textbf{All} represents the entire SpiderUnion dataset.
                The best results for each \#table division are annotated in \textbf{bold}.
            } 
            \label{tab:hop}
        \end{table}
        To verify the effectiveness of multi-hop retrieval in \ourmethod, we compare the performance with varying the number of max hops, as shown in Table~\ref{tab:hop}.
        The results indicate:
        (\emph{i})~\ourmethod performs optimally when the number of max hop is greater than or equal to the number of required tables.
        (\emph{ii})~For questions requiring $1$ or $2$ tables, the best performance is achieved at a max hop of $3$. 
        Since \ourmethod can enhance performance for the questions that retrieve irrelevant tables by guiding to obtain unretrieved relevant tables in subsequent hops.        
        (\emph{iii})~For questions requiring $1$ table, performance slightly decreases because removing retrieved information from such questions could easily introduce errors. 
        (\emph{iv})~The performance of requiring $\geq 4$ tables remains unchanged because improving complete recall $k=5$ requires all relevant tables included in the top $5$ retrieved, which is challenging.

    \subsubsection{Can \ourmethod Reduce the Average Rank of Relevant Tables?}
        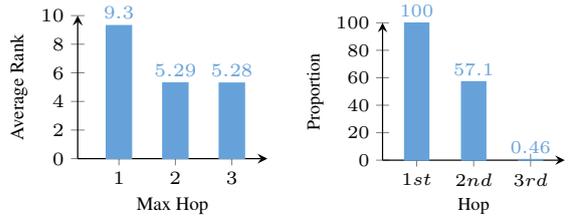
\begin{figure}[t]
            \centering
            \begin{subfigure}[b]{0.48\linewidth}
                \centering
    


        
\resizebox{\linewidth}{!}{
    \begin{tikzpicture}
        \tiny
        \begin{axis}
            [ybar,
             xlabel=Max Hop,
             ylabel=Average Rank,
             xmin=-0.1, xmax=2,
             ymin=0, ymax=10,
             width=\linewidth,
             xtick={0,1,2},
             xticklabels={$1$,$2$,$3$},
             legend style={
                at={(0.90,0.95)}, 
                anchor=north east, 
                font=\large, 
                legend cell align=left
             },
             every node near coord/.append style={
                font=\tiny, 
            },
             bar width=8pt,
             xtick distance=3pt,
             axis y line=left,
             axis x line=bottom,
             nodes near coords,
             enlarge x limits=0.3, 
             nodes near coords={\pgfmathprintnumber[fixed,precision=2]{\pgfplotspointmeta}},
             ] 
             \addplot+ [nodes near coords={\pgfmathprintnumber[fixed,precision=2]{\pgfplotspointmeta}}, cbluee!320] 
             coordinates {
             (0,9.30)
             (1,5.29)
             (2,5.28)
             }; 
        \end{axis} 
    \end{tikzpicture} 
}
            \end{subfigure}
            \hfill 
            \begin{subfigure}[b]{0.48\linewidth}
                \centering
                \resizebox{1\linewidth}{!}{
    \begin{tikzpicture}
        \tiny
        \begin{axis}
            [ybar,
             xlabel=Hop,
             ylabel=Proportion,
             ymin=0, ymax=100,
             xtick={0,1,2},
             width=\linewidth,
             xticklabels={$1st$,$2nd$,$3rd$},
             legend style={
                at={(0.90,0.95)}, 
                anchor=north east, 
                font=\large, 
                legend cell align=left
             },
             every node near coord/.append style={
                font=\tiny, 
            },
             bar width=8pt,
             xtick distance=3pt,
             axis y line=left,
             axis x line=bottom,
             nodes near coords,
             enlarge x limits=0.3, 
             ] 
             \addplot+ [cbluee!320] coordinates {
             (0,100)
             (1,57.1)
             (2,0.46)
             }; 
        \end{axis} 
    \end{tikzpicture} 
}
            \end{subfigure}
            \vspace{-1em}
            \caption{
                The left part is the average rank of relevant tables with different numbers of max hops on the SpiderUnion using \ourmethod.
                The right part is the proportion of questions that are not early stopped with different hops on the SpiderUnion using \ourmethod.
            } 
            \label{fig:rank}
            \vspace{-1em}
        \end{figure}
        To verify \ourmethod can remove the retrieved information from the question and retrieve relevant tables that are not similar to previously retrieved tables, we calculate the average rank of relevant tables at different maximum hops, as shown in the left part of Figure~\ref{fig:rank}. 
        The figure presents that: 
        (\emph{i})~\ourmethod significantly enhances the average rank of relevant tables, with the most notable improvement occurring at a maximum hop of $2$. 
        This is because most questions in SpiderUnion require $1$ or $2$ tables (see Table~\ref{tab:dataset}), requiring two hops to obtain unretrieved relevant tables.
        (\emph{ii})~Conversely, the improvement at a maximum hop of $3$ is weak, as only a small number of questions require $\geq 3$ tables but also because, and most questions complete retrieval before the third hop.

        As illustrated in the right part of Figure~\ref{fig:rank}, the proportion of retrieving at different hops in \ourmethod closely matches the distribution of the number of relevant tables in Table~\ref{tab:dataset}.
        We set the maximum hop to $3$ also because most user questions stop retrieving in the third hop.

    \subsubsection{How Does the Previous Errors Affect Subsequent Performance?}
\begin{figure}
            \centering
    \begin{tikzpicture}
    \small
    \begin{axis}[
        xtick={1,2,3},
        xticklabels={$1st$, $2nd$, $3rd$},
        xlabel style={at={(axis description cs:0.5,-0.1)},anchor=north,font=\small}, 
        xlabel={Hop},
        ylabel={$r@5$},
        ymin=0, ymax=100,
        legend columns=3,
        legend style={
            at={(0.5,-0.3)},
            anchor=north,
            font=\small,
            /tikz/every even column/.append style={column sep=0.3cm}
        },
        nodes near coords,
        point meta=explicit symbolic,
        width=0.8\linewidth,
        grid=both,
        every node near coord/.append style={
                font=\small, 
            },
    ]
    
    \addplot[
        color=corange,
        mark=*,
        mark options={solid},
        style={thick}
    ] table [meta index=2] {
    x y meta
    1 0.0 {0.0}
    2 10.4 {10.4}
    3 11.1 {11.1}
    };
    \addlegendentry{$r@5=0$}
    
    \addplot[
        color=cpurple,
        mark=*,
        mark options={solid},
        style={thick}
    ] table [meta index=2] {
    x y meta
    1 53.3 {53.3}
    2 79.2 {79.2}
    3 78.2 {78.2}
    };
    \addlegendentry{$0<r@5<1$}
    
    \addplot[
        color=cblue,
        mark=*,
        mark options={solid},
        style={thick}
    ] table [meta index=2] {
    x y meta
    1 100.0 {100.0}
    2 97.1 {97.1}
    3 98.0 {98.0}
    };
    \addlegendentry{$r@5=1$}
    
    \end{axis}
    \end{tikzpicture}
            \caption{
                The $r@5$ during hops, categorizing SpiderUnion according to the $r@5$ in the first hop which falls into different intervals.
            }
            \label{fig:error}
        \end{figure}
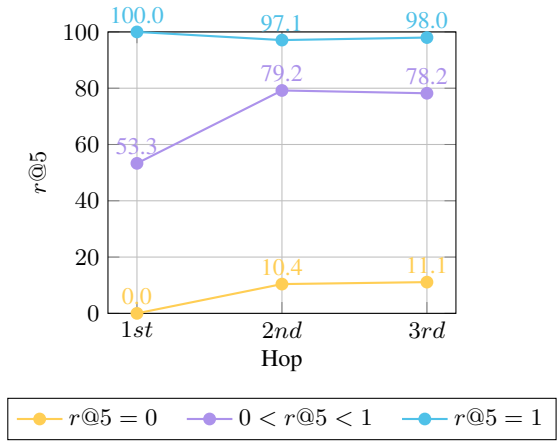

        To analyze the error cascades in \ourmethod, we divide questions based on first-hop results: $r@5=0$, $0<r@5<1$, and $r@5=1$, and compare performance across hops, as shown in Figure~\ref{fig:error}. 
        The error cascades in \ourmethod are minimal and significantly compensated by the performance improvements. 
        
        
        We observe that: 
        (\emph{i})~For questions whose $r@5 = 0$ in the first hop, all the top $5$ tables retrieved are irrelevant. 
        \ourmethod improves performance by removing retrieved information from the questions, thereby eliminating interference from irrelevant tables.
        (\emph{ii})~For questions with $0<r@5<1$ in the first hop, some of the top $5$ tables are irrelevant.
        \ourmethod significantly enhances performance by removing the retrieved information from the user question and focusing on unretrieved information. 
        Performance in the third hop slightly declines mainly because most questions in SpiderUnion involve $\leq 2$ relevant tables, and additional hops may introduce errors. 
        (\emph{iii})~For questions whose $r@5 = 1$ in the first hop, performance in subsequent hops is slightly reduced as all relevant tables are retrieved, causing subsequent hops to introduce minor errors.

\section{Related Work}
    \label{sec:related work}
    \subsection{Text-to-SQL}


    The text-to-SQL task aims to convert user questions into SQL queries, facilitating efficient database access~\cite{qin2022surveytexttosql}.
    LLM-based methods have become mainstream in text-to-SQL due to their superior performance with minimal annotated data~\cite{resdsql,gao2023texttosqldail}.
    For example, \citet{li2024using_rerank} propose using LLMs to predict execution results of test cases, determining the correctness of candidates SQL queries.
    However, most previous methods do not focus on open-domain text-to-SQL, and exist a gap with real-world applications.
    Therefore, CRUSH~\cite{kothyari-etal-2023-crush4sql} uses LLMs to guess potentially relevant tables for retrieval.
    DBCopilot~\cite{wang2024dbcopilot} trains a schema router to identify relevant tables.
    \citet{chen2024table_retrieval} propose a re-ranking relevance method by fine-tuning DTR models.

    However, existing methods are constrained by: \emph{(i)} single-hop retrieval; \emph{(ii)} requiring fine-tuning, which is resource-intensive and domain-specific.
    To solve these problems, we propose a multi-hop table retrieval method for open-domain text-to-SQL.

\subsection{Retrieval-Augmented Generation}
    Retrieval-Augmented Generation (RAG) denotes retrieving information in extra knowledge bases to support accurate and reliable LLM generation \citep{rag_survey_2024}.
    Existing RAG works employ either single-hop retrieval \cite{ma-etal-2023-query-rewriting,zheng2024take-step-back} or multi-hop retrieval, where questions in each hop can be generated in two ways \cite{gao2024rag_survey}: 
    \emph{(i)}~splicing the question with the current and previous retrieval results \cite{shao-etal-2023-enhancing_iterretgen,jeong2024adaptiverag}, or 
    \emph{(ii)}~using LLMs to integrate the retrieved results into the question \cite{press-etal-2023-self_ask,trivedi-etal-2023-ircot,li2023llatrieval}.

    To enhance retrieval in open-domain text-to-SQL, we adopt multi-hop retrieval.
    However, previous methods are unsuitable for this task, because adding retrieved tables to the question could lead to the retrieval of similar yet irrelevant tables, unlike other tasks, where questions require supplementary.
    Therefore, we remove retrieved information from the question for next-hop retrieval.  

\section{Conclusion}
    In the paper, we propose \ourmethod to address the challenge that multi-hop retrieval in other open-domain tasks cannot be directly applied to open-domain text-to-SQL.
    Compared with previous methods, \ourmethod removes the retrieved information from the question to ensure the retrieved tables are relevant to the question.
    Experimental results demonstrate the effectiveness of \ourmethod on two open-domain text-to-SQL datasets.
    Our method achieves new SOTA results compared to previous methods, with an average of $5.7\%$ improvement.

\section*{Limitations}
    We discuss the limitations of our work from the following two aspects.
    (\emph{i})~Considering the applicability, the multi-turn text-to-SQL task is common in real scenarios \cite{yu-etal-2019-cosql,yu-etal-2019-sparc-multi-turn}, while we do not discuss the solutions of open-domain multi-turn text-to-SQL.
    We leave improving our method to apply to multi-turn text-to-SQL for future work.
    (\emph{ii})~From the performance perspective, our method does not consider the performance improvement brought by the text-to-SQL feedback \cite{trivedi-etal-2023-ircot,yu-etal-2023-augmentation_Retriever_Plug-In}. We leave the retrieval recall improvement leveraging the results of text-to-SQL for future work.
    handles ambiguous entities or synonyms within the natural language questions or database schemas.
    Although our method achieves significant improvements, future work can improve our method from the aspects of applicability and recall further.

\section*{Ethics Statement}
    Every dataset and model used in the paper is accessible to the public, and our application of them adheres to their respective licenses and conditions.

\bibliography{custom}
\clearpage

\appendix

\label{sec:appendix}
\section{How to Address the Scenarios Where No Relevant Tables Exist}
\label{appendix:no relevant tables}
    In this section, we discuss how our method can be improved to solve the situation that there are no tables that are relevant to the user question in the given databases \cite{yu-etal-2019-cosql}.
    According to the existing open-domain database tables, we can manually annotate or synthesize relevant and irrelevant questions to train a discriminator.
    For each user question, before retrieving the relevant tables, we can use the discriminator to determine whether the question is irrelevant to the existing tables. 
    And if the question is irrelevant, we can directly output the feedback, and no longer retrieve tables and generate SQL \cite{jeong2024adaptiverag}.

\section{Prompts for Removal}
\label{appendix:prompts for rewrite}
    In the section, we show the prompts we use to remove retrieved information from the question on SpiderUnion (see Table~\ref{tab:prompt_spider}) and BirdUnion (see Table~\ref{tab:prompt_bird}). 
    We also show the prompt without tabulation which is used in \S\ref{subsec:ablation}, as shown in Table~\ref{tab:prompt_spider_without_tabulation}.
    Each table is represented in the form of “database name.table name(column name,
    column name, ...)” following \citet{kothyari-etal-2023-crush4sql}.
    We only show the first two examples here limited by pages.
    The code and the whole prompt will be public in the future. 
    \begin{table*}[t]
        \centering
        \small
        \begin{tabular}{l}
            \toprule
            Given the following SQL tables, your job is to complete the possible left SQL tables given 
            a user’s request.\\ 
            Return None if no left SQL tables according to the user’s request.\\
            \\
            Question: Which models are lighter than 3500 but not built by the 'Ford Motor Company'?\\
            Database: car\_1.model list(model id, maker, model)\\
            car\_1.cars data(id, mpg, cylinders, edispl, horsepower, weight, accelerate, year)\\
            car\_1.car names(make id, model, make)\\
            Completing Tables: car\_1.car makers(maker)\\
            \\
            Question: Which employee received the biggest bonus? Give me the employee name.\\
            Database: employee\_hire\_evaluation.evaluation(employee id, year awarded, bonus)\\ employee\_hire\_evaluation.employee(employee id, name, age, city)\\
            Completing Tables: None\\
            ...\\
            \bottomrule
        \end{tabular}
        \caption{The prompt we use for the SpiderUnion with \texttt{gpt-3.5-turbo}.}
        \label{tab:prompt_spider}
    \end{table*}
    
    \begin{table*}[t]
        \centering
        \small
        \begin{tabular}{l}
            \toprule
            Given the following SQL tables, your job is to complete the possible left SQL tables given 
            a user’s request.\\ 
            Return None if no left SQL tables according to the user’s request.\\
            \\
            Question: What was the growth rate of the total amount of loans across all accounts for a male client between 1996 and 1997?\\
            Database: financial.client(client\_id, gender, birth\_date, location of branch)\\
            financial.loan(loan\_id, account\_id, date, amount, duration, monthly payments, status)\\
            Completing Tables: financial.account(account id, location of branch, frequency, date)\\
            financial.disp(disposition id, client\_id, account\_id, type)\\
            \\
            Question: How many members did attend the event 'Community Theater' in 2019?\\
            Database: student\_club.Attendance(link to event, link to member)\\
            Completing Tables: student\_club.Event(event name, event date)\\
            ...\\
            \bottomrule
        \end{tabular}
        \caption{The prompt we use for the BirdUnion with \texttt{gpt-3.5-turbo}.}
        \label{tab:prompt_bird}
    \end{table*}

    \begin{table*}[t]
        \centering
        \small
        \begin{tabular}{l}
            \toprule
            Remove information appearing in the database from the question .\\
            Return None if the database is totally correspond to the question .\\
            \\
            Question: Which models are lighter than 3500 but not built by the 'Ford Motor Company'?\\
            Database: car\_1.model list(model id, maker, model)\\ car\_1.cars data(id, mpg, cylinders, edispl, horsepower, weight, accelerate, year)\\
            car\_1.car names(make id, model, make)\\
            Rewritten Question: What is the car makers of the 'Ford Motor Company'?\\
            \\
            Question: Which employee received the biggest bonus? Give me the employee name.\\
            Database: employee\_hire\_evaluation.evaluation(employee id, year awarded, bonus)\\ employee\_hire\_evaluation.employee(employee id, name, age, city)\\
            Rewritten Question: None\\
            ...\\
            \bottomrule
        \end{tabular}
        \caption{The prompt we use for the SpiderUnion with \texttt{gpt-3.5-turbo} \textbf{without tabulation}.}
        \label{tab:prompt_spider_without_tabulation}
    \end{table*}

\section{Normalization Method}
\label{appendix:normalization method}
    In this section, we show how to normalize the cosine similarity into a probability distributed between $0$ and $1$ present in Equation~\ref{equ:p_hat}.
    We define the cosine similarity between the question $q$ vector and the table $t_i$ vector as $s$, which is distributed between $-1$ and $1$.
    And we use Equation~\ref{equ:norm} to normalize the cosine similarity $s$.
    \begin{equation}
        Norm(s) = \frac{s+1}{2}
        \label{equ:norm}
    \end{equation}
    Moreover, $Norm(s)$ is proportional to $s$, that is, the greater the cosine similarity $s$, the greater $\hat{P}(t_i | q)$, that is, the greater the probability that the table $t_i$ is retrieved by $q$.

\section{Score of the Retrieval Path}
\label{appendix:score of the retrieval path}
    In this section, we prove the calculation process of the retrieval path probability present in \S\ref{subsec:rank}.
    First of all, we define the retrieval path $\texttt{Path}^{h,b}$ as Equation~\ref{equ:path}, where $q^{h,b}$ represents the user question of hop $h$ and beam $b$, and $t^{q^{h, b}}_{p_h}$ represents the table retrieved by $q^{h, b}$ ranked at $p_h$.
    \begin{equation}
        \texttt{Path}^{h,b} = ((q^{1,b}, t^{q^{1, b}}_{p_1}), ..., (q^{h,b}, t^{q^{h, b}}_{p_h}))
    \label{equ:path}
    \end{equation}
    According to the discussion in \S\ref{subsec:retrieve}, the score of each retrieval path $\texttt{Path}^{h,b}$ can be regarded as the probability that the last table in the path $t^{q^{h, b}}_{p_h}$ is retrieved in the case of the user question at last hop $q^{h, b}$.
    Formally, it can be summarized as:
    \begin{equation}
        \begin{aligned}
            &\texttt{Score\_Path}((q^{1,b}, t^{q^{1, b}}_{p_1}), ..., (q^{h,b}, t^{q^{h, b}}_{p_h}))\\
            =&\hat{P}((q^{1,b}, t^{q^{1, b}}_{p_1}), ..., (q^{h,b}, t^{q^{h, b}}_{p_h}))\\
            =&\hat{P}(t^{q^{h, b}}_{p_h} | q^{h,b}) \cdot \hat{P}((q^{1,b}, t^{q^{1, b}}_{p_1}), ..., (q^{h-1,b}, t^{q^{h-1, b}}_{p_{h-1}}))\\
            =&... \\
            =&\prod_{j=1}^{h}\hat{P}(t^{q^{j, b}}_{p_j} | q^{j,b})
        \end{aligned}
        \label{equ:equ_compute}
    \end{equation}
    Therefore, we multiply all $\hat{P}$ on the retrieval path as the score of the path.

\section{Table Scoring Algorithm in \ourmethod}
\label{appendix:table scoring algorithm}
    In this section, we detail the table scoring algorithm (Algorithm~\ref{alg:algorithm}), which is discussed in \S\ref{subsec:rank}.
    
        \begin{algorithm*}[htbp]
            \begin{algorithmic}[1]
                \item[] {\bfseries Input:}
                    The similarity corresponding to each table $t$ in each hop $h$: $all\_paths=[[(table_{11}, score_{11}), \ldots, (table_{1H}, score_{1H})]$$, \ldots, 
                    [(table_{P1}, score_{P1}), \ldots, 
                    (table_{PH}, $$score_{PH})]]$, 
                    the number of max hops $H$, 
                    the number of all paths $P$.
                \item[] {\bfseries Output:}
                    The scores of each table $t$
                \STATE  Initialization : $table\_score \gets \left\{\right\}$
                \FOR{$each\_path$ in $all\_paths$}
                    \STATE $score \gets 1$
                    \COMMENT{Initialize the score}
                    \FOR{$example$ in $each\_path$}
                        \STATE $score = \adam{score \times example\left[1\right]}$
                        \COMMENT{Calculate the score of the path}
                    \ENDFOR
                    \FOR{$example$ in $each\_path$}
                        \STATE {\begin{tabular}[t]{@{}l@{}}$table\_score\left[example\left[0\right]\right] \gets$$\adam{max(score,$ $table\_score\left[example\left[0\right]\right])}$\end{tabular}} \\
                        \COMMENT{Update the table score with the max path score}
                    \ENDFOR
                \ENDFOR
                \RETURN $table\_score$
            \end{algorithmic}
            \caption{The \adamtext{table scoring} algorithm in \ourmethod}
            \label{alg:algorithm}
        \end{algorithm*}

\section{How to Handle Ambiguous Entities or Synonyms}
\label{appendix:synonyms}
    In this section, we present how our method can be improved to handle ambiguous entities or synonyms within user questions or database tables.
    When scoring the retrieved tables, in the face of multiple similar tables retrieved due to ambiguous entities or synonyms, we can additionally consider the correlation between tables and select tables with higher relevance scores of other retrieved tables \cite{chen2024table_retrieval}.

\section{Dataset Details}
\label{appendix:dataset details}
    In this section, we introduce in detail the source dataset of SpiderUnion and BirdUnion datasets which we use.
    Spider~\cite{yu-etal-2018-spider} is a multi-domain mainstream text-to-SQL dataset that contains $658$ questions, with an average of $1.48$ tables per question in the dev-set.
    Bird~\cite{li2023bird}, as a text-to-SQL dataset, is closer to the actual scenario featuring its larger scale and more difficult questions.
    Bird contains $1534$ questions, with an average of $1.92$ tables per question in the dev-set.

\section{Model Details}
\label{appendix:model details}
    In the section, we introduce the models SGPT and \texttt{gpt-3.5-turbo} used in our experiments.
    SGPT~\cite{muennighoff2022sgpt} is the popular retrieval Single-hop, employing a decoder-only architecture and showing excellent performance on tasks such as sentence matching.
    \texttt{gpt-3.5-turbo}~\cite{zhao2023survey_llm} has undergone instruction fine-tuning and human alignment and has superior in-context learning and inference capability.
    

\section{The Evaluation Metric in Analysis Experiments}
\label{appendix:the evaluation metric used in analysis experiments}
    In this section, we explain the reasons for using complete recall $k=5$ as the evaluation metric in the analysis experiments.
    The increasing trend of the performance in the text-to-SQL becomes slow or even drops when inputting retrieved tables more than $5$ as shown in Table~\ref{tab:text-to-sql}, and considering that SpiderUnion and BirdUnion require up to $4$ tables for each question, so in the following analysis, we are mainly concerned with the performance of the top $5$ retrieval results.
    Furthermore, complete recall $k=5$ is a more strict indicator than $recall@5$, so we mainly utilize complete recall $k=5$ as the evaluation metric in analysis experiments.

\section{Discussion on Efficiency}
\label{appendix:discussion on efficiency}
    In this section, we discuss the comparison of efficiency between \ourmethod and CRUSH.
    Because of each user question, CRUSH needs to use LLM to predict the relevant tables once, and then retrieve all the tables according to the LLM prediction once.
    So the time complexity of CRUSH is shown in Equation~\ref{equ:crush_time}, where $n$ is the number of user questions.
    \begin{equation}
        \begin{aligned}
        T(CRUSH) &= O(2 \cdot n)\\
        &=O(n)
        \end{aligned}
    \label{equ:crush_time}
    \end{equation}
    
    Suppose that the number of hop is $H$ and the beam size is $B$ in \ourmethod.
    For each user question, \ourmethod needs to retrieve all the tables first, and input LLM for removal according to the retrieved top $B$ tables. 
    In the subsequent hops, each hop needs to retrieve $B$ times and remove information $B$ times with LLM.
    Therefore, the time complexity of our method is present in Equation~\ref{equ:murre_time}.
    \begin{equation}
        \begin{aligned}
        T(\ourmethod) &= O((1+B+(H-1) \cdot B \cdot 2) \cdot n)\\
        &= O(B \cdot H \cdot 2 \cdot n) \\
        &= O((B \cdot H) \cdot n)
        \end{aligned}
    \label{equ:murre_time}
    \end{equation}

    It can be found that although our method has significantly improved the performance compared with CRUSH, our method is less efficient.
    However, existing work shows that using reasoning efficiency for improving the reasoning performance has a wide range of practical application value \cite{yao2023treeofthoughts,xie2023selfevaluation,press-etal-2023-self_ask,hashimoto-etal-2024-how-beamsearch}.
    Therefore, in practical applications, how to choose $B$ and $H$ in \ourmethod to achieve a balance between retrieval efficiency and effect should be carefully considered.

\section{Why Retrieve with the Original Question at the First Hop}
\label{app:why not rewrite}
\begin{table*}[ht]
\centering
\small
\begin{tabular}{ll|cccc|cccc}
\toprule
\textbf{Dataset} & \textbf{Method} & \bm{$k=3$} &  \bm{$k=5$} & \bm{$k=10$} &  \bm{$k=20$} & \bm{$r@3$} & \bm{$r@5$} & \bm{$r@10$} & \bm{$r@20$}\\
\midrule
\multirow{2}{*}{BIRDUunion} & \ourmethod & $69.1$ & $80.1$ & $88.7$ & $92.7$ & $81.0$ & $87.6$ & $92.6$ & $95.4$ \\
  & \textit{w tabulation at the first hop} & $52.2$ & $63.6$ & $78.4$ & $88.1$ & $70.0$ & $78.0$ & $87.5$ & $93.0$ \\
\midrule
\multirow{2}{*}{SpiderUnion} & \ourmethod & $86.0$ & $93.5$ & $96.7$ & $97.3$ & $89.3$ & $94.3$ & $96.8$ & $97.5$ \\
  & \textit{w tabulation at the first hop} & $69.3$ & $80.2$ & $88.3$ & $92.2$ & $76.1$ & $85.1$ & $91.1$ & $94.5$ \\
\bottomrule
\end{tabular}
\caption{
The comparison between \ourmethod and the method of tabulationg the question at the first hop, with using SGPT-5.8B as the embedding and using \texttt{gpt-3.5-turbo} to tabulate.
}
\label{tab:rewrite_first_hop}
\end{table*}
To demonstrate the effectiveness of \ourmethod which employs the original user question at the first hop rather than the tabulation question, we conduct experiments, with the results presented in Table~\ref{tab:rewrite_first_hop}.
The results show that starting with tabulation question in table form reduces the retrieval performance, thereby validating the effectiveness of \ourmethod.
The decline in performance is attributed to the absence of the retrieved table to provide domain-constraining information, which increases errors when the question is tabulated at the first hop.

\section{Impact of SQL Hardness}
\label{appendix:impact of sql hardness}
    \begin{table}[htbp]
        \centering
        \small
        \begin{tabular}{l|cccc|c}
            \toprule
            \textbf{Method} & \textbf{Easy} & \textbf{Medium} & \textbf{Hard} & \textbf{Extra} & \textbf{All} \\
            \midrule
            Single-hop & $70.5$ & $71.1$ & $55.8$ & $51.3$ & $66.0$ \\
            \ourmethod & \bm{$71.8$} & \bm{$76.0$} & \bm{$73.3$} & \bm{$73.1$} & \bm{$74.2$ }\\
            \bottomrule
        \end{tabular}
        \caption{Complete recall $k=5$ of \ourmethod compared with the Single-hop in different SQL hardness levels on SpiderUnion. 
        \textbf{Extra} denotes extra hard. 
        \textbf{All} refers to the performance of the whole SpiderUnion dataset.
        The best results of different hardness are annotated in \textbf{bold}.
        }
        \label{tab:level}
    \end{table}
    In this section, we show the performance of \ourmethod on SQL of different hardness levels.
    We categorize the SQL and its corresponding question according to the SQL hardness criteria \cite{yu-etal-2018-spider} and calculate the retrieval performance of different hardness levels, as shown in Table~\ref{tab:level}.
    \ourmethod improves performance more significantly for more difficult SQL questions.
    Because more difficult SQL often requires more tables to operate and query, the Single-hop is challenging to retrieve all relevant tables merely in a single hop, while our method can retrieve more relevant tables with multi-hop retrieval by removing the retrieved information from the question at each hop.

\section{Case Study}
\label{appendix:detailed case study}
    \begin{figure}
        \centering
        \includegraphics[width=1\linewidth]{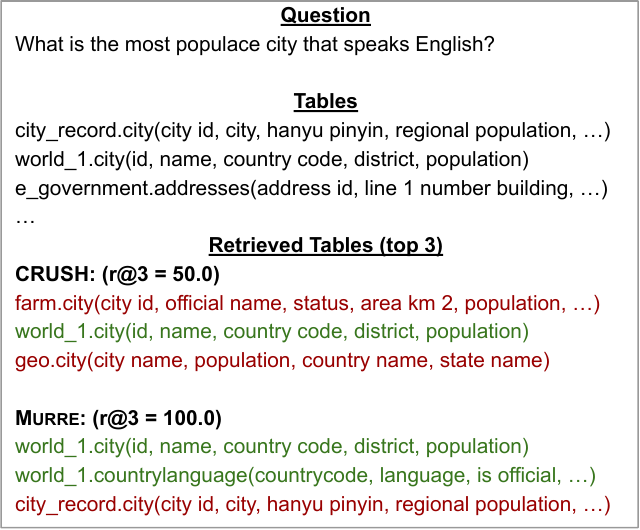}
        \caption{
            A case study comparing \ourmethod with CRUSH. 
            \introgreen{Green} indicates relevant tables, while \introred{red} indicates irrelevant ones. 
            Each table is represented as “database name.table name(column names)”.
            $r$ denotes recall.
        }
        \label{fig:case}
    \end{figure}
    In this section, we compare our methods with other retrieval methods through specific examples to illustrate the advantages of our methods.
    Firstly, we demonstrate a case study with \ourmethod compared with CRUSH, as shown in Figure~\ref{fig:case}.
    We can see that CRUSH fails to retrieve the table \textit{"world\_1.countrylanguage"} within the top $3$ results due to its single-hop retrieval limitation, as the retrieval of table \textit{"world\_1.countrylanguage"} relies on the table \textit{"world\_1.city"}.
    In contrast, \ourmethod employs multi-hop retrieval with a beam size of $3$, increasing the probability of selecting relevant tables at each hop. 
    Additionally, we eliminate the retrieved information in \textit{"world\_1.city"} from the question, which aids the model in retrieving \textit{"world\_1.countrylanguage"} that was previously missed. 
    \begin{table*}[htbp]
        \centering
        \small
        \begin{tabular}{l}
            \toprule
            \textbf{\underline{Question}}\\
            What is the most populace city that speaks English?\\
            \\
            \textbf{\underline{Single-hop ($r@3 = 50.0$)}}\\
            \textbf{Retrieved Tables (top $3$)}\\
            \textcolor{cred}{city\_record.city(city id, city, hanzi, hanyu pinyin, regional population, gdp)}\\
            \textcolor{cgreen}{world\_1.city(id, name, country code, district, population)}\\
            \textcolor{cred}{e\_government.addresses(address id, line 1 number building, town city, zip postcode, state province county, country)}\\
            \\
            \textbf{\underline{\ourmethod ($r@3 = 100.0$)}}\\
            \textbf{Retrieved Tables, hop = $1$ (top $3$)}\\
             \textcolor{cred}{city\_record.city(city id, city, hanzi, hanyu pinyin, regional population, gdp)}\\
            \textcolor{cgreen}{world\_1.city(id, name, country code, district, population)}\\
            \textcolor{cred}{e\_government.addresses(address id, line 1 number building, town city, zip postcode, state province county, country)}\\
            \textbf{Removal Questions}\\
            city\_record.language(city id, language, percentage)\\
            world\_1.countrylanguage(countrycode, language, is official, percentage)\\
            e\_government.languages(language id, language name, language code, population)\\
            \textbf{Retrieved Tables, hop = $2$ (top $3$)}\\
            \textcolor{cgreen}{world\_1.city(id, name, country code, district, population))}\\
            \textcolor{cgreen}{world\_1.countrylanguage(countrycode, language, is official, percentage)}\\
            \textcolor{cred}{city\_record.city(city id, city, hanzi, hanyu pinyin, regional population, gdp)}\\
            \textbf{Removal Questions}\\
            None\\
            None\\
            None\\
            (Early Stop)\\
            
            \bottomrule
        \end{tabular}
        \caption{Detailed case study comparing \ourmethod with Single-hop. The green means the relevant table, while the red means irrelevant. Each table is expressed in the form of “database name.tabel name(column names)”. $r$ denotes recall.}
        \label{tab:detailed case study single-hop}
    \end{table*}

    \begin{table*}[htbp]
        \centering
        \small
        \begin{tabular}{l}
            \toprule
            \textbf{\underline{Question}}\\
            What is the most populace city that speaks English?\\
            \\
            \textbf{\underline{\ourmethod without beam search ($r@3 = 0.0$)}}\\
            \textbf{Retrieved Tables, hop = $1$ (top $3$)}\\
             \textcolor{cred}{city\_record.city(city id, city, hanzi, hanyu pinyin, regional population, gdp)}\\
            \textcolor{cgreen}{world\_1.city(id, name, country code, district, population)}\\
            \textcolor{cred}{e\_government.addresses(address id, line 1 number building, town city, zip postcode, state province county, country)}\\
            \textbf{Removal Question}\\
            city\_record.language(city id, language, percentage)\\
            \textbf{Retrieved Tables, hop = $2$ (top $3$)}\\
            \textcolor{cred}{city\_record.city(city id, city, hanzi, hanyu pinyin, regional population, gdp)}\\
            \textcolor{cred}{city\_record.hosting city(year, match id, host city)}\\
            \textcolor{cred}{city\_record.match(match id, date, venue, score, result, competition)}\\
            \textbf{Removal Question}\\
            None\\
            (Early Stop)\\
            \\
            \textbf{\underline{\ourmethod ($r@3 = 100.0$)}}\\
            \textbf{Retrieved Tables, hop = $1$ (top $3$)}\\
             \textcolor{cred}{city\_record.city(city id, city, hanzi, hanyu pinyin, regional population, gdp)}\\
            \textcolor{cgreen}{world\_1.city(id, name, country code, district, population)}\\
            \textcolor{cred}{e\_government.addresses(address id, line 1 number building, town city, zip postcode, state province county, country)}\\
            \textbf{Removal Questions}\\
            city\_record.language(city id, language, percentage)\\
            world\_1.countrylanguage(countrycode, language, is official, percentage)\\
            e\_government.languages(language id, language name, language code, population)\\
            \textbf{Retrieved Tables, hop = $2$ (top $3$)}\\
            \textcolor{cgreen}{world\_1.city(id, name, country code, district, population))}\\
            \textcolor{cgreen}{world\_1.countrylanguage(countrycode, language, is official, percentage)}\\
            \textcolor{cred}{city\_record.city(city id, city, hanzi, hanyu pinyin, regional population, gdp)}\\
            \textbf{Removal Questions}\\
            None\\
            None\\
            None\\
            (Early Stop)\\
            
            \bottomrule
        \end{tabular}
        \caption{Detailed case study comparing \ourmethod with \ourmethod without beam search. The green means the relevant table, while the red means irrelevant. Each table is expressed in the form of “database name.tabel name(column names)”. $r$ denotes recall.}
        \label{tab:detailed case study wo beam}
    \end{table*}

    \begin{table*}[htbp]
        \centering
        \small
        \begin{tabular}{l}
            \toprule
            \textbf{\underline{Question}}\\
            What is the most populace city that speaks English?\\
            \\
            \textbf{\underline{\ourmethod without Removal ($r@3 = 50.0$)}}\\
            \textbf{Retrieved Tables, hop = $1$ (top $3$)}\\
             \textcolor{cred}{city\_record.city(city id, city, hanzi, hanyu pinyin, regional population, gdp)}\\
            \textcolor{cgreen}{world\_1.city(id, name, country code, district, population)}\\
            \textcolor{cred}{e\_government.addresses(address id, line 1 number building, town city, zip postcode, state province county, country)}\\
            \textbf{Spliced Questions}\\
            What is the most populace city that speaks English?; city\_record.city(city id, city, hanzi, hanyu pinyin,\\regional population, gdp)\\
            What is the most populace city that speaks English?; world\_1.city(id, name, country code, district, population)\\
            What is the most populace city that speaks English?; e\_government.addresses(address id, line 1 number building,\\town city, zip postcode, state province county, country)\\
            \textbf{Retrieved Tables, hop = $2$ (top $3$)}\\
            \textcolor{cred}{city\_record.hosting city(year, match id, host city)}\\
            \textcolor{cred}{county\_public\_safety.city(city id, county id, name, white, black, amerindian, asian, multiracial, hispanic)}\\
            \textcolor{cred}{world\_1.country(code, name, continent, region, surface area, indepdent year, population, life expectancy, gnp, gnp old,}\\
            \textcolor{cred}{local name, ...)}\\
            \textbf{Spliced Questions}\\
            What is the most populace city that speaks English?; world\_1.city(id, name, country code, district, population);\\ city\_record.hosting city(year, match id, host city)\\
            What is the most populace city that speaks English?; world\_1.city(id, name, country code, district, population);\\city\_record.city(city id, city, hanzi, hanyu pinyin, regional population, gdp)\\
            What is the most populace city that speaks English?; city\_record.city(city id, city, hanzi, hanyu pinyin, regional population,\\gdp);\\city\_record.hosting city(year, match id, host city)\\
            \textbf{Retrieved Tables, hop = $3$ (top $3$)}\\
            \textcolor{cred}{city\_record.city(city id, city, hanzi, hanyu pinyin, regional population, gdp)}\\
            \textcolor{cred}{city\_record.hosting city(year, match id, host city)}\\
            \textcolor{cgreen}{world\_1.city(id, name, country code, district, population)}\\
            \\
            \textbf{\underline{\ourmethod ($r@3 = 100.0$)}}\\
            \textbf{Retrieved Tables, hop = $1$ (top $3$)}\\
             \textcolor{cred}{city\_record.city(city id, city, hanzi, hanyu pinyin, regional population, gdp)}\\
            \textcolor{cgreen}{world\_1.city(id, name, country code, district, population)}\\
            \textcolor{cred}{e\_government.addresses(address id, line 1 number building, town city, zip postcode, state province county, country)}\\
            \textbf{Removal Questions}\\
            city\_record.language(city id, language, percentage)\\
            world\_1.countrylanguage(countrycode, language, is official, percentage)\\
            e\_government.languages(language id, language name, language code, population)\\
            \textbf{Retrieved Tables, hop = $2$ (top $3$)}\\
            \textcolor{cgreen}{world\_1.city(id, name, country code, district, population))}\\
            \textcolor{cgreen}{world\_1.countrylanguage(countrycode, language, is official, percentage)}\\
            \textcolor{cred}{city\_record.city(city id, city, hanzi, hanyu pinyin, regional population, gdp)}\\
            \textbf{Removal Questions}\\
            None\\
            None\\
            None\\
            (Early Stop)\\
            
            \bottomrule
        \end{tabular}
        \caption{Detailed case study comparing \ourmethod with \ourmethod without Removal. The green means the relevant table, while the red means irrelevant. Each table is expressed in the form of “database name.tabel name(column names)”. $r$ denotes recall.}
        \label{tab:detailed case study wo rewrite}
    \end{table*}

    We also present one example in detail comparing \ourmethod with the Single-hop, without beam search and without Removal respectively in Table~\ref{tab:detailed case study single-hop}, Table~\ref{tab:detailed case study wo beam}, and Table~\ref{tab:detailed case study wo rewrite}.
    We set the beam\_size to $3$, max hop to $3$, and \ourmethod stops early at the second hop.

    As shown in Table~\ref{tab:detailed case study single-hop}, the Single-hop retrieval fails to retrieve the table \textit{"world\_1.countrylanguage"} at top $3$ limited by the single-hop retrieval since the retrieval of table \textit{"world\_1.countrylanguage"} relies on the table \textit{"world\_1.city"}.
    As displayed in Table~\ref{tab:detailed case study wo beam}, \ourmethod without beam search method is affected by error cascades, because the table \textit{city\_record.city} with the highest retrieval ranking in hop $1$ is irrelevant to the question. 
    Removing based on the irrelevant \textit{city\_record.city} table lead to retrieval errors in subsequent hops.
    As present in Table~\ref{tab:detailed case study wo rewrite}, \ourmethod without Removal adds the retrieved tables directly to the user question, so that the subsequent retrieved tables are similar to the currently retrieved tables.
    For example, the irrelevant table \textit{"city\_record.hosting city"} retrieved in hop $2$ is similar to the table \textit{"city\_record.city"} retrieved in hop $2$, which are both about \textit{"city"} information, but ignore the information of \textit{"language"}.
    Our method focuses on retrieving tables about \textit{"language"} by removing the information \textit{"world\_1.city"} in the retrieved tables, and successfully retrieves two relevant tables.

\section{Statistical Criteria of Limitations}
\label{appendix:statistical criteria of limitations}
    To facilitate statistics on the number of results reflected in the two limitations of Similar Candidate and Semantic Gap, we set the following rules.
    
    For the limitation of Similar Candidate, the statistical standard is that if the irrelevant table that is incorrectly retrieved has the same token as the question, it is considered a Similar Candidate error.
    If the same token appears in the schema as in the question, the cosine similarity between the table and the question after embedding is also high, making it challenging to identify the relevant tables \cite{BM25_Mean,bert_bm25,li2023llatrieval,anonymous2023opentab,wang2024dbcopilot}.
    
    For the limitation of Semantic Gap, the statistical standard is that if the relevant table that has not been retrieved does not have the same token as the question, it is considered a Semantic Gap error.
    If the question does not overlap with any token in the relevant tables, the retrieval similarity is also low, representing the semantic gap to some extent \cite{reimers-gurevych-2019-sentence-bert}.

\end{document}